\pdfoutput=1

\documentclass[11pt]{article}

\usepackage[review]{EMNLP2023}

\usepackage{times}
\usepackage{latexsym}
\usepackage{kotex}  
\usepackage{makecell}   
\usepackage{graphicx}   
\usepackage{anyfontsize}    
\usepackage{graphicx}   

\usepackage{booktabs} 
\usepackage[normalem]{ulem}
\useunder{\uline}{\ul}{} 
\usepackage{lscape}
\usepackage{multirow} 
\useunder{\uline}{\ul}{}
\usepackage{natbib}

\usepackage[T1]{fontenc}

\usepackage[utf8]{inputenc}

\usepackage{inconsolata}

\usepackage{hyperref}   

\title{Improving Korean NLP Tasks with Linguistically Informed Subword Tokenization and Sub-character Decomposition}

\begin{document}
\maketitle
\begin{abstract}
We introduce a morpheme-aware subword tokenization method that utilizes sub-character decomposition to address the challenges of applying Byte Pair Encoding (BPE) to Korean, a language characterized by its rich morphology and unique writing system. Our approach balances linguistic accuracy with computational efficiency in Pre-trained Language Models (PLMs). Our evaluations show that this technique achieves good performances overall, notably improving results in the syntactic task of NIKL-CoLA. This suggests that integrating morpheme type information can enhance language models' syntactic and semantic capabilities, indicating that adopting more linguistic insights can further improve performance beyond standard morphological analysis. Our code is available at \href{https://github.com/taeheejeon22/MorphSubDecomp-Korean}{our GitHub repository}.\footnote{\url{https://github.com/taeheejeon22/MorphSubDecomp-Korean}}

\end{abstract}

\section{Introduction}

Byte Pair Encoding (BPE) \cite{gage1994new, sennrich2015neural} has become the standard tokenization method for Pre-trained Language Models (PLMs), alleviating Out-of-Vocabulary (OOV) issues by breaking words into subwords. Despite being purely data-driven, BPE allows language models possible to utilize linguistic knowledge like compound and derivation. However, BPE's greedy algorithm may yield incorrect or excessive tokenization, often generating morphologically useless subwords. Various studies highlight this shortcoming, pointing out instances such as \textit{chn} in \textit{technology}, and \textit{hell} in \textit{hello}. \cite{aguilar2020char2subword, alkaoud2020importance, bostrom-durrett-2020-byte, park-etal-2020-empirical}.


\begin{figure}
    \centerline{\includegraphics[width=6cm]{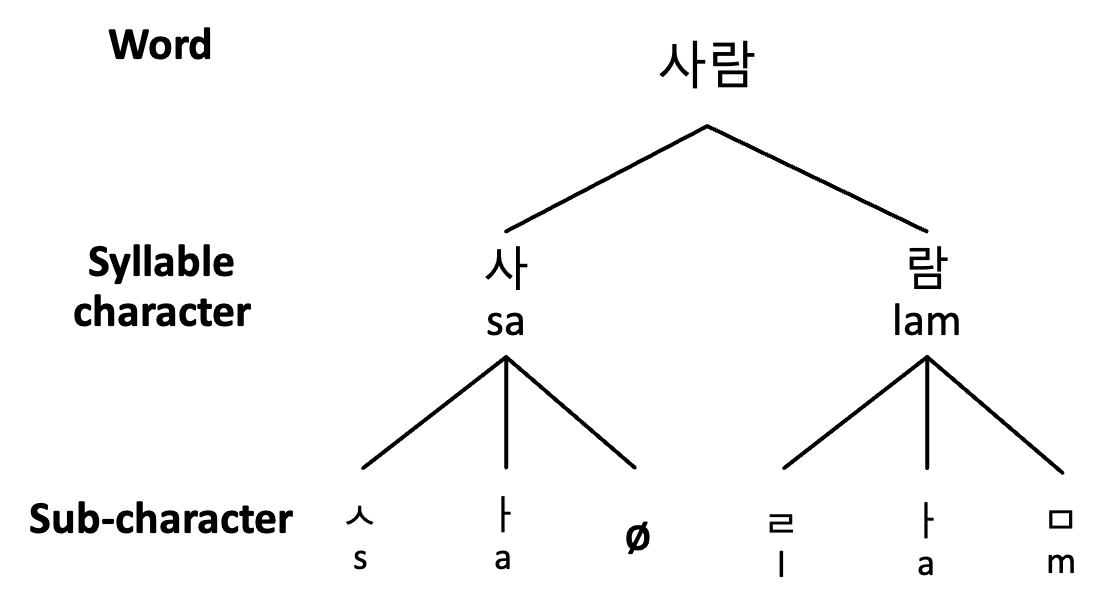}}
    \caption{Two levels of subword information of a Korean word 사람 `human.'}
    \label{fig:1}
\end{figure}
Korean subword tokenization is more complicated because of the syllabic writing system of \textit{Hangul}, the Korean alphabet. First, language models may encounter OOV problems due to infrequent syllable characters. As Figure \ref{fig:1} shows, Korean texts contain two subword information levels, theoretically combining 11,172 syllable-characters with sub-characters. While frequent syllable-characters are limited because of the orthography, syllable-characters which occur in typos, misspellings, and neologisms can cause issues. Second and more importantly, syllable-character-level subword tokenization would not fully capture Korean's agglutinative grammar. A single syllable character can be related to multiple morphemes, both lexical and grammatical. For example, sub-character ㄹ in a syllable 갈 `will go`, represent a future tense morpheme. Ignoring sub-character-level subword information of Korean can lead to the failure of solving OOV problems and losing crucial linguistic information.
 

Two main solutions have been used to address these problems. First, sub-character decomposition simply breaks syllable-characters into sub-characters, capturing crucial subword information (e.g., -ㄹ \textit{future tense} from 갈 `will go'). This is common in studies using traditional word embeddings \cite{park-etal-2018-subword, kim-etal-2022-break}. Second, morpheme analysis, by employing a morphological analyzer, enable the models to detect subword morphemes directly, regardless of subword levels. For instance, 갈 'will go' can be segmented into two morphemes 가- 'go' and -ㄹ 'will'. This method is popular not only for traditional word embeddings but also for PLMs \cite{park-etal-2020-empirical, kim-etal-2021-changes}.



While many studies have addressed these issues, there is a pressing need for further research particularly for PLMs using BPE-style subword tokenization. Sub-character decomposition in PLM subword tokenizations, though potentially leading to meaningless subwords (e.g., ㄺㅇㅡㄴ from 해맑은 `bright'), has not been explored sufficiently. Only a few studies \cite{lee2020kr, kim2021enhancing} investigated this method on PLMs. Furthermore, the efficacy of sub-character decomposition with morpheme-aware subword tokenization \cite{park-etal-2020-empirical, kim-etal-2021-changes}, which have improved Korean tokenization successfully, should be validated. Although some studies \cite{Youn2020wordvec, jeon2022tokft} have addressed this method using traditional word embeddings, to our knowledge, it remains untested on PLMs.
In this regard, we propose a linguistically-informed BPE-based subword tokenization for Korean NLP tasks, focusing on morpheme type information. To evaluate the efficacy of our method, we adopt a pre-training fine-tuning approach using various Korean NLP tasks. We selected the Bidirectional Encoder Representations from Transformers (BERT) \cite{devlin2018bert}, considering its proven effectiveness in language comprehension and reasonable training cost.

%



\section{Related Work}

    \subsection{Subword Tokenization with Sub-character Decomposition}
Since BPE-based subword tokenization methods, such as Wordpiece \cite{wu2016google}, and Sentencepiece \cite{kudo2018subword} alleviate the OOV problem by segmenting words into subwords, they became standard tokenization methods for various languages. 

As a language-specific subword tokenization methods for Korean, sub-character decomposition has been proposed.
\citet{park-etal-2018-subword} and \citet{lee2020kr} reported improved performance of Korean NLP tasks utilizing traditional word embedding and PLMs respectively by leveraging sub-character-level tokenization. \citet{kim-etal-2022-break} decomposes sub-character into sub-components of sub-character for Korean word embedding. These fine-grained decomposition approaches, while robust against typos and neologisms, increase computational complexity due to excessively long sequences.



    \subsection{Morpheme-aware Subword Tokenization}
Recently, hybrid approaches aiming to capture the morphological characteristics of Korean have gained attention. \citet{park-etal-2020-empirical} reported that BERT models utilizing a combination of morpheme analysis and BPE-based subword tokenization outperformed BERT models using only BPE-based subword tokenization or sub-character tokenization. Studies like \citet{park-etal-2020-empirical} and HyperCLOVA \citep{kim-etal-2021-changes} have shown improved performance by utilizing morhpeme-aware subword tokenization compared to using BPE alone. 

Techniques utilizing morpheme analysis alongside sub-character decomposition have also been introduced. \citet{Youn2020wordvec} proposed a method that decomposes words into sub-character at the morpheme level with part-of-speech (POS) information. \citet{jeon2022tokft} adopted the method of \citet{Youn2020wordvec}, and additionally suggested selective sub-character decomposition according to the type of morpheme, whether lexical or grammatical.






\section{Linguistically Informed Subword Tokenization and Sub-character Decomposition}






    \subsection{Challenges of Korean Writing System in Morpheme Type Consideration} \label{3.1.}
In Korean linguistics, morphemes are generally classified as two types, lexical and grammatical. Lexical morphemes such as nouns and verbs have their own lexical meanings. Grammatical morphemes represent grammatical relationships among lexical morphemes. Particles (case makers and postpositions) and endings (inflectional suffixes) are typical grammatical morphemes of Korean.  

\begin{figure}
    \centerline{\includegraphics[width=8cm]{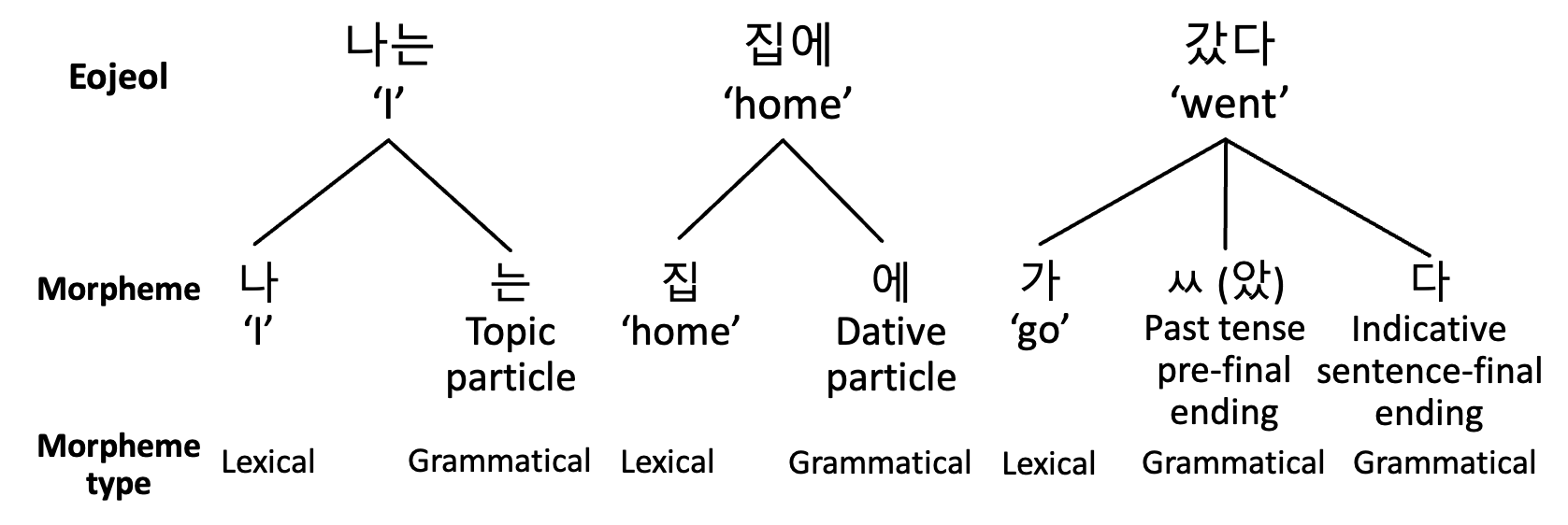}}
    \caption{Morphological analysis with morpheme type information of a Korean sentence 나는 집에 갔다 `I went home.'}
    \label{fig:2}
\end{figure}

It is important that there are many grammatical morphemes which do not form full syllable-characters. Multiple morphemes including both lexical and grammatical morphemes can occur in a syllable in Korean due to its syllabic writing system. In Figure \ref{fig:2}, the eojeols, Korean spacing units, which consist of syllable-characters are divided into multiple morphemes. It is noted that the decomposition of syllable-characters into sub-characters is necessary to obtain certain grammatical morphemes which do not form full syllable-characters like ㅆ (originally \textit{past tense ending} -았-) in 갔다 `went.' Thus, to equip a language model with comprehensive linguistic information of grammatical morphemes, we should utilize sub-character-level subword information.


    \subsection{Morpheme Tokenization with Sub-character Decomposition} \label{3.2}
With regard to Korean subword tokenization, both sub-character decomposition \cite{lee2020kr} and morpheme tokenization \cite{park-etal-2020-empirical, kim-etal-2021-changes} can be used to capture the linguistic information of grammatical morphemes. Each approach has its pros and cons. Sub-character decomposition, unlike morpheme tokenization does not need morphological analyzers which can cause analysis errors. On the other hand, morpheme tokenization can suppress the generation of unwanted meaningless subwords due to pre-trained morphological analyzers. Then, what if we first perform morpheme tokenization, and perform sub-character decomposition subsequently?



Using both together seems to be unnecessary because each can capture sub-character-level subword information independently. Nevertheless, sub-character decomposition after morpheme tokenization may have certain advantages. As morphological analyzers would potentially cause errors, some morphemes could be not fully, or wrongly segmented especially in ambiguous strings. For example, an eojeol 나는 can be analyzed either 나 `I' + 는 (nominative particle) or 날- `fly' + -는 (adnominal ending). In addition, morphological analyzers would struggle with typos, misspellings (e.g., 재밋다 as a typo of 재밌다 `interesting') and neologisms absent in their training data. Sub-character decomposition after morpheme tokenization is expected to alleviate such problems.




\begin{table*}[htb!]
\centering
\begin{tabular}{ll}
\Xhline{3\arrayrulewidth}
\textbf{\small Tokenization} & \textbf{\small Tokenized Sequence} \\
\Xhline{3\arrayrulewidth}
{\small Raw Text} & {\small 나\underline{라면} 해물\underline{라면}을 먹었을걸.} \\ 
\hline

{\small WP} & {\small 나라/ \#\#면/ 해/ \#\#물/ \#\#라면/ \#\#을/ 먹었/ \#\#을/ \#\#걸 \#\#.} \\

{\small WP-SD}  & {\small ㄴㅏㄹㅏ/ \#\#ㅁㅕㄴ/ ㅎㅐ/ \#\#ㅁㅜㄹ/ \#\#ㄹㅏㅁㅕㄴ/ \#\#ㅇㅡㄹ/ ㅁㅓㄱㅇㅓㅆㅇㅡㄹ/ \#\#걸.} \\

{\small MorWP} & {\small 나/ 이/ 라면/ 해물/ 라면/ 을/ 먹/ 었/ 을걸/ . }\\

{\small MorWP-SD} & {\small ㄴㅏ/ ㅇㅣ/ \textit{\underline{ㄹㅏㅁㅕㄴ}}/ ㅎㅐㅁㅜㄹ/  \underline{ㄹㅏㅁㅕㄴ}/ ㅇㅡㄹ/ ㅁㅓㄱ/ ㅇㅓㅆ/ ㅇㅡㄹㄱㅓㄹ/ .} \\

{\small MorWP-MD} & {\small ㄴㅏ/ 이/ \textbf{\underline{라면}}/ ㅎㅐㅁㅜㄹ/ \underline{ㄹㅏㅁㅕㄴ}/ 을/  ㅁㅓㄱ/ 었/ 을걸/ .} \\

\Xhline{3\arrayrulewidth}
\end{tabular}
\caption{Tokenization examples of a Korean sentence 나라면 해물라면을 먹었을걸. `If I had been you, I would have eaten seafood ramen.' Wordpiece prefixes (\#\#) are added to the `pieces' of full tokens (eojeols or morphemes). Note that the subword 라면 can be either a lexical morpheme (a noun) `ramen' or a grammatical morpheme (a connective ending). MorWP-MD does not apply sub-character decomposition to the first subword 라면 which is a grammatical morpheme.}
\label{tab:2}
\end{table*}

    \subsection{Why not Simple Sub-character Decomposition?} \label{3.3}

Sub-character decomposition may not always be beneficial for all pre-tokenized tokens in BPE-style subword tokenization. This view supported by \citet{jeon2022tokft}, which demonstrated selective sub-character decomposition—applied to either lexical or grammatical morphemes—outperforms simple decomposition in the context of traditional word embeddings. However, this raises the question: which morphemes should be decomposed?

It appears that we can benefit from sub-character decomposition on lexical morphemes rather than grammatical ones. Basically, most of the segmented sub-character strings of grammatical morphemes (e.g., ㄴㅡ from \textit{topic particle} 는) are totally meaningless in Korean. In addition, the OOV problem of grammatical morphemes can hardly occur given grammatical morpheme's significantly higher usage frequency compared to lexical morphemes.


 
On the other hand, lexical morphemes are unlimited in number. No matter how much data we use, the OOV problem of lexical morphemes will occur inevitably. It could be caused by lack of data, vocabulary size limitation, misspellings, typos, etc. Sub-character decomposition of lexical morpheme tokens which are pre-tokenized by a morphological analyzer is expected to alleviate these types of problems.





    \subsection{Morpheme-aware Subword Tokenization with Morphologicial Sub-character Decomposition} \label{3.4}

In this paper, we propose a hybrid subword tokenization method for Korean PLMs which integrates morpheme type information. Basically, we adopt morpheme-aware subword tokenization following \citet{park-etal-2020-empirical}, but we introduce an additional step prior to BPE modelling. As discussed in Section \ref{3.3}, we implement sub-character decomposition only for lexical morphemes.

During training, PLMs are expected to benefit from morphological sub-character decomposition. Compared to static word embeddings like Word2Vec \cite{mikolov2013distributed}, PLMs such as BERT can deal with polysemy (e.g., 사람: `a human being', `the personality of a person', etc.) effectively due to their contextualized word representation. Although these models can deal with homonymy (e.g., 눈: `snow', `eye', etc.) effectively, homonymy may present greater challenges than polysemy. Specifically, unlike polysemes, homographs whose POSs are different from each other (e.g., 어서: \emph{adverb} `quickly' or \emph{connective ending}) can occur in totally different linguistic environments. This could complicate the contextualization task for PLMs.

Morphological sub-character decomposition can enable PLMs to identify morpheme types, which is part of the POS information. Assuming no morpheme analysis errors, lexical morphemes will almost always consist of sub-characters, and vice versa. Consequently, PLMs can capture a broader range of linguistic information without requiring additional labels.

\section{Tokenization Methods}
For the comparison purpose, we use five tokenization methods utilizing morpheme tokenization, BPE, and sub-character decomposition. Table \ref{tab:2} shows the examples of each method.

    \subsection{Subword Tokenization (WP)}
For subword tokenization, we use WordPiece \cite{wu2016google}, which is based on the BPE algorithm. It tokenizes every eojeol into subwords. HuggingFace's Tokenizers\footnote{\url{https://huggingface.co/docs/transformers/main_classes/tokenizer}}
 is used as the implementation of the WordPiece tokenizer.

    \subsection{Subword Tokenization with Simple Sub-character Decomposition (WP-SD)}
We perform simple sub-character decomposition for the training data before training a WordPiece model. Using this method, not all, but many grammatical morphemes can be segmented as subwords. This will make language models utilize more grammatical information. In addition, sub-character decomposition can help to solve OOV problems and incorrect spacing problems according to \citet{lee2020kr}, which tested this method on BERT models.

    \subsection{Morpheme-aware Subword Tokenization (MorWP)}
Following \citet{park-etal-2020-empirical}, we first implement morpheme tokenization using a morphological analyzer, and then train a WordPiece model. This method can not only capture morphological information of Korean effectively, but also prevent a WordPiece tokenizer from generating unwanted subwords.
    \subsection{Morpheme-aware Subword Tokenization with Simple Sub-character Decomposition (MorWP-SD)}
Before WordPiece modeling, we apply simple sub-character decomposition on the morpheme-tokenized text. As mentioned in Section \ref{3.2}, the OOV problem can be alleviated further than MorWP.

    \subsection{Morpheme-aware Subword Tokenization with Morphological Sub-character Decomposition (MorWP-MD)}

MorWP-SD can be enhanced by morphological sub-character decomposition, as discussed in Section \ref{3.4}. To be specific, we first perform morpheme tokenization using a morphological analyzer. In sequence, we apply sub-character decomposition only to lexical morphemes using POS tags provided by morphological analyzer, then we train a WordPiece model.

\section{Experiments}

\subsection{Pre-training}

    \subsubsection{Dataset}
For pre-training BERT models, we use the recent dump of Korean Wiki (780MB) \footnote{https://dumps.wikimedia.org/kowiki/latest/} and Namuwiki (5.5GB) \footnote{https://dump.thewiki.kr}. We extract the plain text from them using WikiExtractor \footnote{https://github.com/attardi/wikiextractor} and Namu Wiki Extractor \footnote{https://github.com/jonghwanhyeon/namu-wiki-extractor} respectively. We remove characters which are not ASCII characters or Hanguls. We also remove sentences which are shorter than 3 eojeols. For sub-character decomposition, we utilize Unicode Normalization Form Canonical Decomposition.

    \subsubsection{Morpheme Tokenization}
For morpheme tokenization, we use MeCab-ko\footnote{https://bitbucket.org/eunjeon/mecab-ko}, one of the most stable morphological analyzers for Korean. Based on MeCab \cite{kudo2006mecab} which is a Japanese morphological analyzer, MeCab-ko has been widely-used for many Korean NLP tasks \cite{kim-etal-2019-segmentation, park-etal-2020-empirical, park-etal-2021-find, lee-shin-2021-korean, kim2021enhancing}.


We use KoNLPy \cite{park2014konlpy}, one of the most popular Korean NLP library, for the MeCab-ko Python API. However, it does not provide completely segmented results for the eojoels which have syllables concerned with multiple morphemes. For example, an eojeol 갔다 `went' which consists of three morphemes (가- `go', -았- past tense ending, 다 indicative ending) is segmented into 갔 + 다. For our purpose, the syllable 갔 should be segmented further into 가 (가- `go') + ㅆ (-았- past tense ending). To solve this problem, we utilize a fixed version of the API\footnote{https://github.com/taeheejeon22/LinguisticTokenization-Korean-ft} from \citet{jeon2022tokft}.

    \subsubsection{WordPiece Modeling}
We train 10 tokenizers, which combines the various versions of tokenization methods mentioned above with WordPiece. For each tokenizer, the vocabulary size is set to 32K or 64K, and other hyperparameters are set to the default values. The Korean Wiki and Namuwiki are used as the dataset for WordPiece training.
    
    \subsubsection{BERT Pre-training}
We use the official BERT pre-train code\footnote{https://github.com/google-research/bert} to train BERT-Base models. The hyperparameters are as follows: batch size = 1024, warm-up step = 10K, learning rate = 5e-5, max sequence length = 128. These hyperparameters are all in accordance with \citet{park-etal-2020-empirical}. Other hyperparameters are as follwos: `--do\_lower\_case' == False, duplicate factor = 5, optimizer = AdamW \cite{loshchilov2017decoupled}. We utilize the Google cloud TPU v3-8 for pre-training, and it takes about 4 to 5 days per each model.


\begin{table}
\centering
\begin{tabular}{lcc}
\hline
\Xhline{3\arrayrulewidth}

\multicolumn{1}{c}{\multirow{2}{*}{\textbf{\small Task}}} & \multicolumn{2}{c}{\textbf{\small Vocab Size}}            \\ \cline{2-3} 
\multicolumn{1}{c}{}                               & \multicolumn{1}{c}{{\small 32K}}  & {\small 64K}  \\ \hline
\Xhline{3\arrayrulewidth}

{\footnotesize NIKL-CoLA}                                          & \multicolumn{1}{c}{\footnotesize 64 / 1e-5 / 3} & {\footnotesize 64 / 1e-5 / 3} \\ \hline
{\footnotesize KLUE-DP}                                             & \multicolumn{1}{c}{\footnotesize 32 / 5e-5 / 9} & {\footnotesize 32 / 5e-5 / 10} \\ \hline
{\footnotesize NSMC}                                                & \multicolumn{1}{c}{\footnotesize 64 / 2e-5 / 3} & {\footnotesize 64 / 2e-5 / 2} \\ \hline
{\footnotesize HSD}                                                 & \multicolumn{1}{c}{\footnotesize 64 / 5e-5 / 5} & {\footnotesize 32 / 5e-5 / 4} \\ \hline
{\footnotesize KLUE-NLI}                                            & \multicolumn{1}{c}{\footnotesize 64 / 5e-5 / 5} & {\footnotesize 32 / 3e-5 / 5} \\ \hline
{\footnotesize PAWS-X}                                                & \multicolumn{1}{c}{\footnotesize 32 / 5e-5 / 5} & {\footnotesize 64 / 5e-5 / 5} \\ \hline

\Xhline{3\arrayrulewidth}

\end{tabular}

\caption{Fine-tuning hyper-parameters for tasks. Hyperparameters are expressed as batch size / learning rate /
epoch.}
\label{tab:3}
\end{table}

\begin{table*}[ht!]
\resizebox{\textwidth}{!}{%
\begin{tabular}{clcccccccccccc}
\Xhline{3\arrayrulewidth}
\hline
\multicolumn{1}{l}{\multirow{3}{*}{\textbf{Vocab Size}}} & \multirow{3}{*}{\textbf{Tokenization}} & \multicolumn{4}{c}{\textbf{Syntax}}                                                       & \multicolumn{3}{c}{\textbf{Semantics}}                             & \multicolumn{3}{c}{\textbf{Syntax \& Semantics}}                              & \multirow{3}{*}{\textbf{\begin{tabular}[c]{@{}c@{}}OOV \\ Rate (\%)\end{tabular}}} & \multirow{3}{*}{\textbf{\begin{tabular}[c]{@{}c@{}}Wordpiece \\ Subtoken \\ Rate (\%)\end{tabular}}} \\ \cline{3-12}
\multicolumn{1}{l}{}                                     &                                        & \multicolumn{2}{c}{\textbf{NIKL-CoLA}}      & \multicolumn{2}{c}{\textbf{KLUE-DP}}        & \multicolumn{2}{c}{\textbf{NSMC}}           & \textbf{HSD}         & \textbf{KLUE-NLI}    & \multicolumn{2}{c}{\textbf{PAWS-X}}                    &                                                                                    &                                                                                                   \\ \cline{3-12}
\multicolumn{1}{l}{}                                     &                                        & Dev                  & Test                 & UAS                  & LAS                  & Dev                  & Test                 & Dev                  & Dev                  & Dev                             & Test                 &                                                                                    &                                                                                                   \\ \hline
\multirow{5}{*}{32K}                                     & WP                                     & 57.62                & 61.64                & 92.56                & 87.02                & 90.06                & 89.52                & 64.82                & 76.20                & \multicolumn{1}{r}{76.60}       & 72.39                & 0.78                                                                               & 54.63                                                                                             \\
                                                         & WP-SD                                  & 59.61                & 59.85                & 92.58                & 87.21                & 89.69                & 89.38                & 64.09                & 76.23                & \multicolumn{1}{r}{78.20}       & 75.23                & 0.57                                                                               & 51.42                                                                                             \\
                                                         & MorWP                                  & 63.62                & {\ul 67.87}          & 92.55                & 87.15                & 90.65                & 90.11                & 65.81                & 76.55                & \multicolumn{1}{r}{77.90}       & 73.99                & 0.68                                                                               & 12.79                                                                                             \\
                                                         & MorWP-SD                               & 64.79                & 67.34                & 92.63                & 87.30                & {\ul 90.92}          & 90.20                & {\ul 66.67}          & {\ul 76.85}          & \multicolumn{1}{r}{78.57}       & 75.12                & 0.47                                                                               & 10.43                                                                                             \\
                                                         & MorWP-MD                               & {\ul 65.19}          & 67.21                & {\ul 92.63}          & {\ul \textbf{87.30}} & 90.84                & {\ul 90.24}          & 65.46                & 76.83                & \multicolumn{1}{r}{{\ul 79.37}} & {\ul 75.27}          & 0.69                                                                               & 10.37                                                                                             \\ \hline
\multirow{5}{*}{64K}                                     & WP                                     & 59.15                & 60.21                & 92.65                & 87.06                & 89.73                & 89.53                & 61.98                & 76.99                & 78.14                           & 73.57                & 0.90                                                                               & 47.96                                                                                             \\
                                                         & WP-SD                                  & 58.76                & 60.91                & {\ul \textbf{92.88}} & 87.12                & 89.82                & 89.63                & 62.20                & 76.72                & 79.33                           & 74.49                & 0.63                                                                               & 46.58                                                                                             \\
                                                         & MorWP                                  & 64.66                & 67.47                & 92.74                & {\ul 87.29}          & 90.82                & {\ul \textbf{90.40}} & 66.07                & 76.84                & {\ul \textbf{79.76}}            & 75.88                & 0.72                                                                               & 7.55                                                                                              \\
                                                         & MorWP-SD                               & 65.54                & 67.09                & 92.38                & 87.28                & {\ul \textbf{90.96}} & 90.38                & {\ul \textbf{68.55}} & 76.90                & 79.61                           & 75.57                & 0.49                                                                               & 6.98                                                                                              \\
                                                         & MorWP-MD                               & {\ul \textbf{66.32}} & {\ul \textbf{69.64}} & 92.84                & 87.27                & 90.95                & 90.39                & 66.62                & {\ul \textbf{78.01}} & 79.42                           & {\ul \textbf{76.22}} & 0.72                                                                               & 6.88                                                                                              \\ \hline
\end{tabular}%
}
\caption{Performance of various models on several NLP tasks and OOV rate, Wordpiece Subtoken Rate (WSR) of each model. The best scores in each column (global) are bold-faced and underlined, and the best scores in each column (local) are underlined. The metrics of each task are as follows: NIKL-CoLA: Accuracy, KLUE-DP: Macro F1 (UAS, LAS), NSMC: Accuracy, HSD: Macro F1, KLUE-NLI: Accuracy, PAWS-X: Accuracy. And OOV Rate, WSR are caculated as: OOV Rate = The number of OOV tokens / The number of tokens * 100, WSR = The number of Wordpiece subtokens / The number of tokens * 100}
\label{tab:4}
\end{table*}

\subsection{Fine-tuning}
With the BERT models trained using each tokenizer, we conduct fine-tuning on several tasks to validate our proposal. All experiments are conducted with Pytorch and Huggingface Transformers \citep{wolf-etal-2020-transformers}. We modify and use the code\footnote{https://github.com/KLUE-benchmark/KLUE-baseline} of \citet{park2021klue} for KLUE-NLI, KLUE-DP. For the other tasks, we modify and use the code\footnote{https://github.com/kakaobrain/kortok} of \citet{park-etal-2020-empirical}.
In order to find the optimal hyperparameters for fine-tuning, we select the batch size from \{32, 64\}, the learning rate from \{1e-5, 2e-5, 3e-5, 5e-5\}, and the epoch from 1 to 10. We experiment with these settings with 5 random seeds, and the hyperparameters with the highest average dev set score for 5 random seeds are finally selected. The finally selected hyperparameters are presented in Table \ref{tab:3}. In the case of KLUE-DP, hyparameters with the highest average of LAS and UAS are selected. We use the AdamW as an optimizer. All fine-tuning tasks were performed utilizing four RTX 2080 Ti graphics cards, with each fine-tuning task allocated a single RTX 2080 Ti for execution.

    \subsubsection{Dataset}
We use 6 datasets for fine-tuning. The types of dataset and tasks are divided into syntactic tasks (NIKL-CoLA, KLUE-DP), semantic tasks (NSMC, HSD), and complex (syntactic \& semantic) tasks (KLUE-NLI, PAWS-X).

\textbf{NIKL-CoLA (The Corpus of Linguistic Acceptability of MODU corpus)}\footnote{https://corpus.korean.go.kr/} is similar to The Corpus of Linguistic Acceptability in GLUE \citep{wang2018glue}. The sentences of NIKL-CoLA were extracted from 1,851 journal articles on linguistic theory. The train set consists of 15,876 sentences, the dev set consists of 1,060 sentences, and the test set consists of 972 sentences. We only use the in-domain set. The label of the dataset consists of two classes: acceptable, and not acceptable.

\textbf{KLUE-DP} \citep{park2021klue} is a dataset for dependency parsing task.
The training set of KLUE-DP consists of 10,000 sentences, and the dev set consists of 2,000 sentences. Since the test set is not disclosed, we use only the dev set for evaluation. We use unlabeled attachment score (UAS) and labeled attachment score (LAS) as the metrics following \citet{park2021klue}. UAS is a macro F1 score for head prediction, and LAS is a macro F1 score for dependency label prediction, calculated only if the head prediction is correct.

\textbf{Naver Sentiment Movie Corpus (NSMC)}\footnote{https://github.com/e9t/nsmc} is a dataset for sentiment analysis task. The original dataset consists of 150,000 trainsets and 50,000 test sets, but 10\% of the train set is used as the dev set following \citet{park-etal-2020-empirical}. Accordingly, the train set has a total of 135,000 sentences, the dev set has 15,000 sentences, and the test set has 50,000 sentences. The label of the dataset consists of two classes: positive, and negative.

\textbf{Korean Hate Speech Dataset (HSD)} \citep{moon2020beep} is a dataset for the Hate Speech Detection task. It contains sentences collected from comments on entertainment news. The label of the dataset consists of three classes: hate, offensive, and neutral. The train set consists of 7,896 sentences, the dev set consists of 471 sentences, and the test set is not disclosed.

\textbf{KLUE-NLI} \citep{park2021klue} is the task that infers semantic relationships between hypothesis and premise sentences. The label of the dataset consists of three classes: entailment, contradiction, and neutral. The train set consists of 24,998 sentence pairs and the dev set consists of 3,000 sentence pairs. The test set is not disclosed.

\textbf{PAWS-X} \citep{yang2019paws}  is a paraphrase identification dataset. We use only the Korean dataset among these. The train set consists of 49,401 sentence pairs, the dev set consists of 1,965 sentence pairs and the test set consists of 1,972 sentence pairs. The label of the dataset consists of two classes: same meaning, and not same meaning.


\begin{figure*}[ht]
    \centerline{\includegraphics[width=\textwidth]{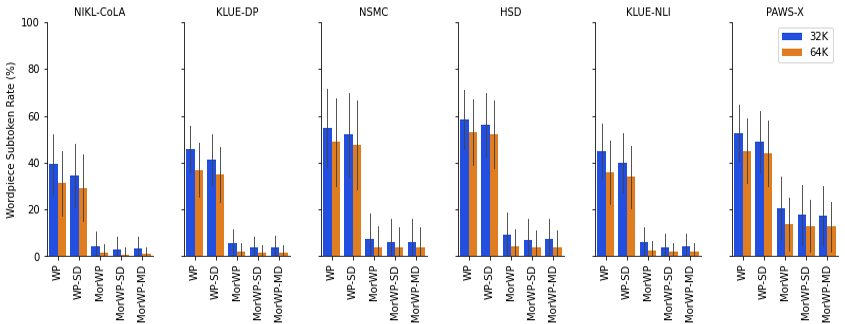}}
    \caption{WSR per sentence (\%) by dataset and tokenizer. From the left, WSR bar plot of NIKL-CoLA, KULE-DP, NSMC, HSD, KLUE-NLI, and PAWS-X are represented. The blue bars represent 32K models, and the orange bars represent 64K models. The error bar drawn in the center of the bar represents the standard deviation.} 
    \label{fig:3}
\end{figure*}




    \subsection{Results}


Table \ref{tab:4} shows the performance of models. The main points of the results are as follows. First, the larger the vocabulary size, the better overall performance. This is in line with the results of \citet{park-etal-2020-empirical}. Second, The MorWP-MD models and The MorWP-SD models achieve the best scores for the majority of the tasks in both 32K versions and 64K versions. Third, the task with the most distinct performance difference for each model is NIKL-CoLA. In particular, it is noteworthy that the performance of the MorWP-MD-64k model is more than 3.2\% ahead of the MorWP-64k model, which is second in performance on test sets. Fourth, in the case of KLUE-DP, the differences between the models are very small compare to the other tasks. The standard deviations of KLUE-DP are only 0.04 (UAS) and 0.12 (LAS) for the 32K versions, and 0.20 (UAS) and 0.10 (LAS) for the 64K versions. Fifth, in HSD, the MorWP-SD-64K model scores the highest, followed by the MorWP-MD-64K model.

Meanwhile, we can see that the OOV rates of all models are less than 1\%. In contrast, the WSRs (Wordpiece Subtoken Rates) have a lot of differences between the models. The WSRs of the WP models and the WP-SD models are higher than 46\%, but the WSRs of the MorWP, the MorWP-SD, and the MorWP-MD are less than 13\%.



\section{Discussion}

    \subsection{OOV and Wordpiece Subtoken Rates}
We find that the performance of BERT models is significantly influenced by the WSRs (the rates of \#\# token) than the OOV rates. As demonstrated in Table \ref{tab:4}, the OOV rates (the rates of UNK token) seem quite low regardless of the method. However, the WSR differences among the methods are much greater. These differences become even more apparent in Figure \ref{fig:3}, which shows the WSR per sentence by dataset and tokenizer. Considering the results in Table \ref{tab:4}, the BERT model's performance improve as the WSR decreases in general.

The correlation between WSR and PLM performance can also be explained in terms of the WSER (Wordpiece Subtoken Entry Rate) of BERT vocabulary. From Table \ref{tab:5}, we first observe that tokenization methods utilizing morpheme tokenization (MorWP, MorWP-SD, and MorWP-MD), which show better performance in Table \ref{tab:4}, have considerably lower WSRs than the others. Secondly, we also find that the WSERs of the 64K tokenization methods are lower than their counterparts. Given that these methods incorporate more full eojeol or morpheme entries, which can be trained more easily than Wordpiece entries, the 64K models turn out to perform better than the 32K models on the whole as shown in Table \ref{tab:4}. These results suggest that the fewer Wordpiece entries the vocabulary of a PLM has, the better the performance of the model will be. Although tokenizers must deal with OOV problems at all costs, the use of Wordpiece tokens should be considered as the \textit{last resort}.
\begin{table}[ht!]
\centering
\begin{tabular}{lcc}
\Xhline{3\arrayrulewidth}
\hline
\multicolumn{3}{c}{\textbf{\begin{tabular}[c]{@{}c@{}}\small Wordpiece Subtoken \\ \small Entry Rate (WSER) (\%)\end{tabular}}}                                    \\ \hline
\multicolumn{1}{c}{\multirow{2}{*}{\textbf{{\small Tokenization}}}} & \multicolumn{2}{c}{\textbf{\begin{tabular}[c]{@{}c@{}}\small Vocabulary \\ \small Size\end{tabular}}} \\
\multicolumn{1}{c}{}                                       & {\small 32K}                   & {\small 64K}                  \\ \hline
{\small WP}                                                         & {\small 29.50}                 & {\small 25.03}                \\ \hline
{\small WP-SD}                                                      & {\small 28.08}                 & {\small 23.70}                \\ \hline
{\small MorWP}                                                      & {\small 16.34}                 & {\small 13.83}                \\ \hline
{\small MorWP-SD}                                                   & {\small 15.62}                 & {\small 13.62}                \\ \hline
{\small MorWP-MD}                                                   & {\small 15.96}                 & {\small 13.76}                \\ \hline
\Xhline{3\arrayrulewidth}
\end{tabular}
\caption{WSERs of each BERT vocabulary. This shows how many partial tokens which are pieces of eojoels or morphemes (e.g., \#\#들에게 from an eojeol 사람들에게 `to people') are included as vocabulary entries in each vocabulary.}
\label{tab:5}
\end{table}

    \subsection{Effect of Morpholgical Sub-character Decomposition} \label{6.2}
To verify the effectiveness of our language-specific approach, we focus on MorWP-SD, and MorWP-MD which implement sub-character decomposition after morpheme tokenization. As seen in Figure \ref{fig:3}, roughly speaking, the lower the WSR, the higher the performance of PLMs tends to be. However, although the WSR per sentence of the MorWP-SD models are lower than that of the MorWP-MD models, the MorWP-MD models outperform the MorWP-SD models in the majority of the tasks as seen in Table \ref{tab:4}. Considering that the OOV problem of grammatical morphemes hardly occur, there will be no significant difference between their performances of solving the OOV problem. Then, what makes the difference?

The superiority of the MorWP-MD models, which utilize morphological sub-character decomposition, would be attributed to their quality of contextualization on homographic vocabulary entries which either can be lexical morphemes or grammatical morphemes. For example, as seen in Table \ref{tab:2}, a morpheme token 라면 can be interpreted as a noun 라면 `ramen' or a subjunctive ending -라면 according to the contexts. As discussed in Section \ref{3.4}, the MorMD-based models have the advantage of contextualizing those tokens.

\begin{table*}[htb!]
\centering
\begin{tabular}{ll}
\Xhline{3\arrayrulewidth}
\textbf{Raw Text} & {\small 그는 \textbf{우크라이나} 사람이다.} \\
\hline

\textbf{Correct Morpheme Analysis} & {\small 그/NP 는/JX \underline{우크라이나/NNP} 사람/NNG 이/VCP 다/EF ./SF} \\
\hline

\textbf{Erroneous Analysis of MeCab-ko} & {\small 그/NP 는/JX \underline{우크라/NNP 이나/JC} 사람/NNG 이/VCP 다/EF ./SF} \\
\hline

\textbf{MorWP-MD Tokenization} & {\small ㄱㅡ/ 는/ \underline{ㅇㅜㅋㅡㄹㅏ/ 이나/} ㅅㅏㄹㅏㅁ/ 이/ 다/ .} \\

\Xhline{3\arrayrulewidth}
\end{tabular}
\caption{An example of a morpheme analysis error leading to tokenization failure. The raw text means `He is Ukrainian.' Note that although 우크라이나 `Ukraine' is a single noun, MeCab-ko analyzes it as a concatenation of two morphemes, 우크라 (an abbreviation of 우크라이나 `Ukraine') and 이나 (auxiliary postpositional particle). MorWP-MD does not apply sub-character decomposition to 이나 which is analyzed as a grammatical morpheme by MeCab-ko, resulting in failure of MorWP-MD Tokenization. The lables /NP, /NNP, /NNG, /JX, /JC,  /VCP, /EF, and /SF correspond to the POS tags for pronoun, proper noun, common noun, auxiliary particle, conjunctive particle,  copula, and sentence-ending punctuation, respectively}
\label{tab:6}
\end{table*}

    \subsection{Analysis of the fine-tuning tasks}

Overall, the MorWP-MD models perform better in syntactic tasks than in semantic tasks. The 64K MorWP-MD model performs best in the NIKL-CoLA task, which is mainly concerned with syntax. The MorWP-MD models also perform well in the KLUE-NLI and the PAWS-X tasks, where both syntactic and semantic information is important. As discussed in Section \ref{6.2}, our method effectively handles homographic grammatical morphemes, indicating potential advantages in these types of tasks.

In general, the MorWP-SD models outperform the MorWP-MD models in semantic tasks. Since lexical morphemes such as nouns and adjectives are critical to classifying the sentiment, syntactic information of grammatical morphemes appears not to play a significant role in these tasks. Moreover, comment datasets like NSMC and HSD often contain neologisms, typos, and spacing errors. To address these issues, splitting words into sub-characters can be beneficial. Additionally, MorWP-MD models occasionally tokenize proper nouns inappropriately. For example, MorWP-MD models split proper nouns like '박유천' to '박유 \#\#천' or '유인나' to '유인 \#\#나'. It can be speculated that because of the separate storage of grammatical morphemes as distinct entries in the MorWP-MD models, certain infrequent words such as proper nouns may not be included in the vocabulary.


In the case of KLUE-DP, the performance differences between all models in both the 32K and the 64K models are less than 1.0. Even the performance differences between the WP models and the other models were insignificant. Similarly, the DP task showed minimal performance differences among models in the original KLUE paper by \cite{park2021klue}. We speculate that the DP task might be challenging for current PLMs, potentially diluting the effectiveness of different tokenization methods. 

\section{Conclusion}
In this study, we explored subword tokenization for Korean, particularly examining sub-character decomposition after morpheme tokenization within BPE frameworks. Our results indicate that this technique not only mitigates errors from morphological analyzers but also improves the BERT models' interpretation of homographs. The incorporation of morphological sub-character decomposition significantly improves performance, notably in contextualizing homographs and in syntactic tasks, with the best results using a 64K vocabulary size.

Due to their language-independent property, BPE-based tokenizers have solved OOV problems successfully regardless of language. However, once a BPE-based tokenizer achieves a certain degree of tokenization quality, solving OOV problems with subwords seems to have little influence on the performance of the whole PLM. Our study has shed light on how PLMs comprehend linguistic information and how their syntactic and semantic performance can be influenced by access to morpheme type information. Our findings suggest that adopting broader linguistic insights beyond mere morphological data could further refine PLM performance.

\section{Limitation}
The most significant limitation of this study is that the proposed method, morphological sub-character decomposition, is highly dependent on the performance of a morphological analyzer. Unfortunately, no Korean morphological analyzer currently available is error-free. MeCab-ko, one of the most stable Korean morphological analyzers, is no exception. The errors of a morphological analyzer can lead to failures of not only morpheme tokenization, but also morphological sub-character decomposition. Table \ref{tab:6} shows the example of this type of error. In this example, 이나 (auxiliary particle) which is part of a full lexical morpheme 우크라이나 `Ukraine' is not decomposed as sub-characters properly. Such tokenization errors will adversely affect not only the training of the lexical morpheme 우크라이나 and the whole sentence but also the contextualization of the grammatical morpheme 이나.

We attribute the slight superiority of the MorWP-MD models over the MorWP-SD models to these errors. However, viewed from another perspective, the MorWP-MD models outperformed the MorWP-SD models despite these errors. Therefore, it is important to develop more stable morphological analyzer to enhance the quality of Korean tokenization further.

\section*{Acknowledgements}
For pre-training models, Cloud TPUs from the TensorFlow Research Cloud program were employed. This work was supported by the KIST Institutional Program (2E32282) and by the National Research Council of Science Technology (NST) grant by the Korea government (MSIP) (No. CAP21051-200 and CRC-20-04-KIST).


\bibliography{anthology,custom}
\bibliographystyle{acl_natbib}

\appendix



\end{document}